\documentclass[conference]{IEEEtran}
\IEEEoverridecommandlockouts

\usepackage{cite}
\usepackage{amsmath,amssymb,amsfonts}
\usepackage{algorithm}
\usepackage{algpseudocode}
\usepackage{graphicx}
\usepackage{textcomp}
\usepackage{xcolor}
\def\BibTeX{{\rm B\kern-.05em{\sc i\kern-.025em b}\kern-.08em
    T\kern-.1667em\lower.7ex\hbox{E}\kern-.125emX}}
\usepackage{booktabs}
\usepackage{float}
\usepackage{subcaption}
\usepackage{multirow}
\usepackage{hyperref}
\usepackage{fancyhdr}

\begin{document}

\fancypagestyle{ieee-copyright}{
    \fancyhf{} 
    \renewcommand{\headrulewidth}{0pt} 
    \fancyfoot[C]{ 
        \footnotesize 
        \parbox{\textwidth}{ \centering 
            \copyright~2025 IEEE. Personal use of this material is permitted. Permission from IEEE must be obtained for all other uses, in any current or future media, including reprinting/republishing this material for advertising or promotional purposes, creating new collective works, for resale or redistribution to servers or lists, or reuse of any copyrighted component of this work in other works.
            \\ \vspace{4pt}
            Accepted for publication in the IEEE International Conference on Tools with Artificial Intelligence (ICTAI), 2025.
            \\
            Final version available at: \url{https://doi.ieeecomputersociety.org/10.1109/ICTAI66417.2025.00061}
        }
    }
}

\title{Integer Linear Programming Preprocessing for Maximum Satisfiability\\
}


\author{\IEEEauthorblockN{1\textsuperscript{st} Jialu Zhang}
\IEEEauthorblockA{\textit{Laboratoire MIS UR 4290} \\
\textit{Université de Picardie Jules Verne}\\
Amiens, France \\
jialu.zhang@u-picardie.fr}
\and
\IEEEauthorblockN{2\textsuperscript{nd} Chu-Min Li}
\IEEEauthorblockA{\textit{Laboratoire MIS UR 4290} \\
\textit{Université de Picardie Jules Verne} \\
Amiens, France \\
chu-min.li@u-picardie.fr}
\and
\IEEEauthorblockN{3\textsuperscript{rd} Sami Cherif}
\IEEEauthorblockA{\textit{Laboratoire MIS UR 4290} \\
\textit{Université de Picardie Jules Verne}\\
Amiens, France \\
sami.cherif@u-picardie.fr}
\and
\IEEEauthorblockN{4\textsuperscript{th} Shuolin Li*}\thanks{*Corresponding author}
\IEEEauthorblockA{\textit{CNRS, LIS} \\
\textit{Aix Marseille University,}
Marseille, France \\
shuolin.li@lis-lab.fr}
\and
\IEEEauthorblockN{5\textsuperscript{th} Zhifei Zheng}
\IEEEauthorblockA{\textit{Laboratoire MIS UR 4290} \\
\textit{Université de Picardie Jules Verne,}
Amiens, France \\
zhifei.zheng@u-picardie.fr}
}

\maketitle
\thispagestyle{ieee-copyright}

\begin{abstract}
The Maximum Satisfiability problem (MaxSAT) is a major optimization challenge with numerous practical applications. In recent MaxSAT evaluations, most MaxSAT solvers have incorporated an Integer Linear Programming (ILP) solver into their portfolios. However, a good portfolio strategy requires a lot of tuning work and is limited to the profiling benchmark. This paper proposes a methodology to fully integrate ILP preprocessing techniques into the MaxSAT solving pipeline and investigates the impact on the top-performing MaxSAT solvers. Experimental results show that our approach helps to improve 5 out of 6 state-of-the-art MaxSAT solvers, especially for WMaxCDCL-OpenWbo1200, the winner of the MaxSAT evaluation 2024 on the unweighted track, which is able to solve 15 additional instances using our methodology.
\end{abstract}

\begin{IEEEkeywords}
MaxSAT, ILP, Preprocessing.
\end{IEEEkeywords}

\section{Introduction}
\label{sec:Introduction}

Maximum Satisfiability (MaxSAT) is a natural optimization extension of the Propositional Satisfiability problem (SAT).
While SAT consists in determining an assignment that satisfies the clausal constraints in a given formula under Conjunctive Normal Form (CNF), the goal in MaxSAT shifts to finding a solution satisfying the maximum number of clauses in the formula, which is more challenging~\cite{handbookofSAT_chapter24_Fahiem,li:hal-04321437}.
Many optimization problems can be formulated as MaxSAT instances, including scheduling~\cite{DBLP:conf/cp/CherifSLLD24,MaxSATforTimetabling,DBLP:conf/ictai/ZhengCS24,aaai19Schedul}, hardware and software debugging~\cite{Chen2010AutomatedDD,SOCdesignApplication}, explainable artificial intelligence~\cite{maxsat_for_ai_hao,MaxSAT_AI_Alexey}, MaxClique~\cite{incMCQ}, Graph coloring~\cite{gcplearning}, and Maximum Commun Subgraph problem~\cite{aaai20MCS}, among many others.

Algorithms for solving the MaxSAT problem can be broadly classified into exact algorithms and heuristic algorithms. Exact algorithms, such as SAT-based (e.g., RC2~\cite{ignatiev_rc2_2019}), Branch and Bound (e.g., MaxCDCL~\cite{li_combining_2021} and WMaxCDCL~\cite{wmaxcdcl_2025_jordi,wmaxcdcl_2025_li}), Integer Linear Programming (ILP), and Tableau calculus~\cite{tableauMaxSAT2016}, find the optimal solution and prove its optimality. In contrast, heuristic algorithms, such as stochastic local search~\cite{bandmaxsat}, can also be competitive, but they do not guarantee optimality. It is known that ILP solvers, while they perform well on certain families of instances, are not competitive for most industrial and random instances~\cite{Anstegui2013SolvingP}. Therefore, the common practice observed in recent MaxSAT evaluations is to combine ILP solvers in a portfolio with other types of solvers to solve MaxSAT instances. For example, in the MaxSAT evaluation 2024~\cite{maxsat_evaluation_2024}, the total time limit to solve an instance is 3600s; EvalMaxSAT~\cite{EvalMaxSAT_2024_Florent} first runs the ILP solver SCIP \cite{achterberg_scip_2009} for 400s and then itself for 3200s; UWrMaxSat~\cite{UWrMaxSAT_2024_Marek} runs SCIP and itself alternatively, each with a possibly different time limit. As such, all these portfolio MaxSAT solvers use an ILP solver independently, requiring careful heuristic tuning, such as setting specific time limits.

In this paper, we propose an integrated approach that incorporates an ILP solver as a preprocessing step in the solving pipeline. Our approach consists in first reading the CNF formula to convert it using integer linear constraints with an objective function, then the ILP solver is called to simplify the problem, and finally, the simplified integer linear constraints are re-encoded into CNF to be solved using a MaxSAT solver. Experimental results show that this approach allows state-of-the-art MaxSAT solvers to solve more instances. Furthermore, we note that our approach does not require setting a heuristic time limit for ILP solvers, allowing them to run until they fully preprocess the instance, which is different from other approaches, such as EvalMaxSAT and UWrMaxSat.

The remainder of this paper is organized as follows. Section~\ref{prelim} introduces the MaxSAT problem and reviews preprocessing techniques. Section \ref{meth} presents the methodology for integrating ILP-based preprocessing techniques into MaxSAT solvers. Section \ref{Exp} discusses our experimental results. Finally, we conclude and discuss future work in Section~\ref{Conc}.

\section{Preliminaries} \label{prelim}

\subsection{Maximum Satisfiability}

Given a set of Boolean variables, a literal $l$ is either a variable $x$ or its negation $\neg x$. A clause $c$ is a disjunction of literals and can be represented as a set of literals. A formula $F$ in Conjunctive Normal Form (CNF) is a conjunction of clauses. A variable $x$ is assigned if it takes a value in $\{True, False\}$ (i.e., $\{1,0\}$). A literal $x$ ($\neg x$) is assigned to $True$ ($False$) if variable $x$ is set to $True$, and to $False$ ($True$) otherwise. A clause $c$ is satisfied if at least one of its literals is assigned to $True$. A formula $F$ is satisfied if all its clauses are satisfied. The Satisfiability (SAT) problem consists in finding an assignment that satisfies a given CNF formula $F$.

MaxSAT is the natural optimization extension of SAT\footnote{Other optimization extensions include, for instance, Minimum Satisfiability (MinSAT) \cite{minsat}}, encompassing both Partial MaxSAT and Weighted Partial MaxSAT~\cite{handbookofSAT_chapter24_Fahiem, li:hal-04321437}. Partial MaxSAT divides clauses into hard clauses $H$ and soft clauses $S$, i.e., $F{=}H{\cup} S$, and the goal is to find an assignment that satisfies all hard clauses in $H$ while maximizing the number of satisfied soft clauses in $S$. In Weighted Partial MaxSAT, a soft clause $c{\in}S$ can be falsified with an integer penalty $w_c$, also called the weight of $c$. The objective for Weighted Partial MaxSAT is thus to find an optimal assignment that maximizes (minimizes) the sum of weights of satisfied (falsified) soft clauses while satisfying all the hard clauses.

\subsection{Preprocessing Techniques}

Preprocessing in problem-solving usually amounts to transforming a given instance into an equivalent one that would potentially be easier to solve. In ILP solvers, preprocessing techniques are a key factor to speed up problem-solving, including variable fixing, variable aggregation, redundant constraint elimination, and other advanced inference mechanisms~\cite{savelsbergh_preprocessing_1994}. The variable fixing technique employs a probing algorithm that temporarily assigns a binary variable to 0 or 1 and then propagates the resulting implications~\cite{achterberg_scip_2009}. Variable aggregation exploits equations and constraint relationships within the model, as well as cluster or symmetry detection algorithms, to merge multiple variables into a single one. Meanwhile, redundant constraint elimination checks the bounds of each constraint, removes constraints that are proved to be satisfied by all variable values satisfying other constraints, or detects constraints implying infeasibility of the problem~\cite{ILP_presolve_Achterberg}.

In SAT and MaxSAT solvers, preprocessing techniques are broadly used to reduce variables and clauses, such as bounded variable elimination, failed literal detection, unit propagation, and self-subsuming resolution~\cite{MaxSAT_preprocessing_Argelich,SAT_preprocessing_Armin,esa_2025_li}. Clause vivification~\cite{AIJchumin,vivification_2017_luo} can also be used as preprocessing or inprocessing among hard clauses. Most of these techniques are incorporated in state-of-the-art solvers, but there are also independent tools that implement SAT and MaxSAT preprocessing techniques, such as MaxPre~\cite{gaspers_maxpre_2017} and Coprocessor~\cite{Coprocessor2}.

A MaxSAT problem can be naturally converted into an ILP problem, as showcased below. Equations \eqref{eq:ILPModel_1}-\eqref{eq:ILPModel_5} give an ILP model for the weighted partial MaxSAT problem $F=H{\cup}S$, where $H$ ($S$) is the set of hard (soft) clauses, respectively. $V$ is the set of Boolean decision variables in $F$. A binary variable $z_c$ is introduced for each soft clause $c\in S$ with weight $w_c$, a binary variable $y_x$ is introduced for each Boolean variable $x$ in $V$, and a hard (soft) clause $c$ is written as  $H_c^- \lor H_c^+$ ($S_c^- \lor S_c^+$), where $H_c^-$ ($H_c^+$) is a disjunction of negative (positive) literals. Equation \eqref{eq:ILPModel_2} ensures that every hard clause is satisfied, and Equation \eqref{eq:ILPModel_3} entails that if a soft clause $c$ is satisfied, then its weight $w_c$ can contribute to the objective function.
However, the encoding of an ILP problem into MaxSAT is not so straightforward, as many ILP constraints require sophisticated techniques to be efficiently encoded into MaxSAT. Fortunately, tools such as the PBLib Library \cite{PBLib_Springer_2015} provide efficient encoding techniques.

\begin{subequations}    
\label{eq:ILPModel}
  \begin{align}
     & \textbf{Objective:} \quad
    \text{Maximize } \sum\limits_{c \in S} w_c \cdot z_c                                                      \label{eq:ILPModel_1}   \\
     & \textbf{Subject to:} \nonumber                                                                                                    \\
     & \sum\limits_{x \in H_c^+} y_x + \sum\limits_{x \in H_c^-} (1-y_x) \geq 1, \quad \forall c \in H \quad  \label{eq:ILPModel_2}   \\
     & z_c \leq \sum\limits_{x \in S_c^+} y_x + \sum\limits_{x \in S_c^-} (1-y_x), \quad \forall c \in S \quad  \label{eq:ILPModel_3} \\
     & z_c \in \{0, 1\}, \quad \forall c \in S                                                                  \label{eq:ILPModel_4}    \\
     & y_x \in \{0, 1\}, \quad \forall x \in V     \label{eq:ILPModel_5}
  \end{align}
\end{subequations}

\section{Methodology} \label{meth}

We propose a methodology to integrate ILP preprocessing techniques into MaxSAT solvers. This section presents our methodology as well as the variable and constraint encodings.

\subsection{Integrating ILP Preprocessing into MaxSAT Solving}
\label{ThreeStagesMethod}

Our three-stage methodology can be described as follows:

\begin{enumerate}
    \item \textbf{Preprocessing Stage:} Given a MaxSAT instance ($originInst$), an ILP model ($originModel$) is constructed based on Equations \eqref{eq:ILPModel_1} to \eqref{eq:ILPModel_5}. Preprocessing techniques are then applied to $originModel$ using an ILP solver, yielding a hopefully simplified model ($simpModel$). The $simpModel$ is subsequently encoded into a simplified MaxSAT instance ($simpInst$), while the mapping between variables in $originInst$ and $simpInst$ is recorded in $varMap$.
    
    \item \textbf{Solving Stage:} If $simpInst$ is ``smaller'' than $originInst$ (i.e., if $simpInst$ contains fewer variables {\em and} fewer hard clauses than $originInst$), a MaxSAT solver is applied to solve $simpInst$ to obtain an optimal solution ($simpSol$) of $simpInst$. Otherwise, the original instance ($originInst$) is solved by the MaxSAT solver. The definition of ``smaller'' is based on our experimental data (see Section~\ref{ILP_statistic}), as the percentage of fixed and aggregated variables in ``smaller'' instances is higher than in other instances.
    
    \item \textbf{Reconstruction Stage:} In this stage, the algorithm constructs an optimal solution $originSol$ for $originInst$ with $simpSol$ and $varMap$. This stage happens only when $simpInst$ is ``smaller'' than $originInst$. 
\end{enumerate}

\begin{figure}[t]
  \centering
  \includegraphics[width=0.5\textwidth]{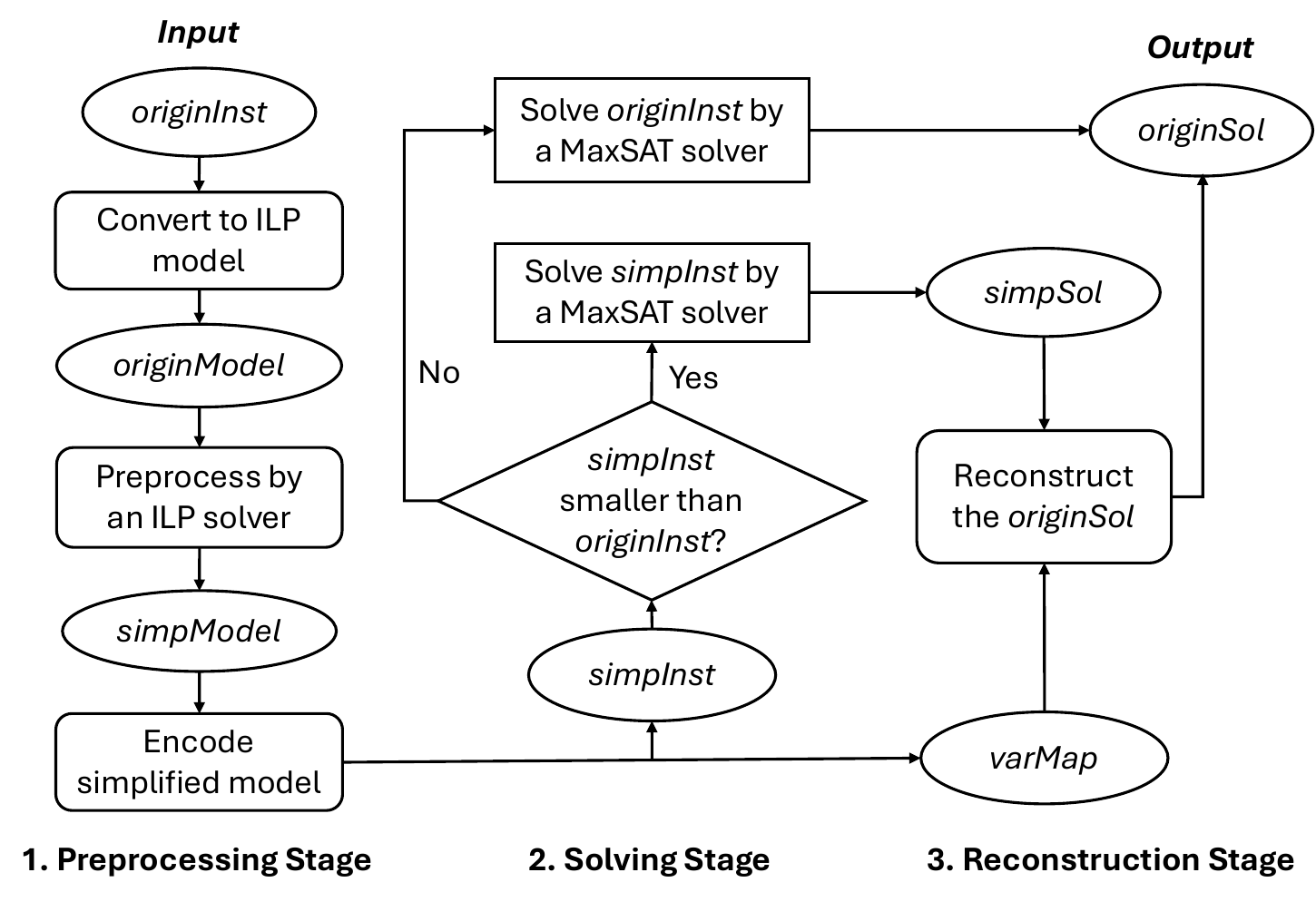}
  \caption{Pipeline of the three-stage methodology}
  \label{fig:methodology_v1}
\end{figure}

The three stages are illustrated in Figure~\ref{fig:methodology_v1}. The key aspect is to convert $simpModel$ into $simpInst$. We first check the variables and constraints in $simpModel$ and then try to encode them to MaxSAT. The encoding involves mapping variables from $simpModel$ to $simpInst$, encoding constraints as hard clauses, and representing the objective function as soft clauses. The details are described in the following subsections.

\begin{table*}[!h]
    \centering
    \caption{Encodings of different constraints in $simpModel$}
    \label{tab:constraints_encoding}

    \renewcommand{\arraystretch}{1.5} 
    \begin{tabular}{|c|c|c|} \hline 
         Constraint&  Formula&Encoding\\ \hline 
         Logical OR& 
     $\sum_{i=1}^{n} x_i \geq 1$&$(x_1 \vee x_2 \vee ... \vee x_n)$\\ \hline
 Logical AND& $\prod_{i=1}^{n} x_i = y$&$(y \vee \neg x_1 \vee ... \vee \neg x_n) \land \bigwedge_{i=1}^{n} (\neg y \vee x_i) $\\\hline
 Setppc packing& $\sum_{i=1}^{n} x_i \leq 1$ &at-most-one\\\hline
 Setppc partitioning&$\sum_{i=1}^{n} x_i = 1$&at-most-one $\land  (x_1 \vee x_2 \vee ... \vee x_n)$ \\\hline
 Linear& $lhs \leq \sum_{i=1}^{n} {w_i \cdot x_i} \leq rhs$&pseudo-Boolean\\\hline\end{tabular}
\end{table*}

\subsection{Variable Encoding}

After preprocessing by an ILP solver, the decision variables in the $originModel$ are transformed into the $simpModel$, in which we distinguish three types of binary variables and introduce them as follows:

\begin{itemize}
    \item \textbf{Fixed}: A variable is considered fixed in $simpModel$ if it is assigned a fixed value, as assigning it the opposite value is provably inconsistent, i.e., it would falsify at least one constraint. Algorithm~\ref{alg:encode_variables} records the values of fixed variables in $varMap$ (line \ref{algo:fixedVar}) for the reconstruction of $originSol$.

    \item \textbf{Aggregated}: A variable $y_x$ in $simpModel$ is referred to as \emph{aggregated} if it can be expressed as a linear combination of other variables $y_1, \dots, y_n$, i.e.,
\[
y_x = c_0 + \sum_{i=1}^{n} c_i \cdot y_i,
\]
where $c_0, c_1, \dots, c_n$ are constants. This entails that the value of $y_x$ depends on other variables $y_i$ for $i=1, \ldots,n$. A special case of aggregation, called \emph{simple aggregation}, occurs when $n=1$ and the coefficients take the form $c_0{=}0, c_1{=}1$, giving $y_x{=}y_1$, or $c_0{=}1, c_1{=}-1$, giving $y_x{=}1-y_1$. Thus, in simple aggregation, $y_x$ either directly mirrors $y_1$ or represents its negation, respectively.
    When such aggregations are encountered, Algorithm~\ref{alg:encode_variables} traverses the aggregation chain and creates a unique new Boolean variable to represent all variables in the chain by preserving their relations (lines \ref{algo:simpAgg1}-\ref{algo:simpAgg2}). In the general case, Algorithm~\ref{alg:encode_variables} encodes the aggregation constraint as a pseudo-Boolean formula 
    and translates it into hard clauses in $simpInst$ (lines \ref{newVarForAgg}-\ref{pbForAgg}).
    \item \textbf{Free}: a variable $y_x$ is referred to as free if it is neither fixed nor aggregated. Algorithm~\ref{alg:encode_variables} creates a new Boolean variable in $simpInst$ for each free variable in $simpModel$ (line \ref{createNewVar}).
\end{itemize}

\begin{algorithm}[!t]
  \caption{Encoding Variables}
  \label{alg:encode_variables}
  \begin{algorithmic}[1]
    \Require $originInst$, $simpModel$, $varMap$

    \For {each variable $x$ \textbf{in} $originInst$}
        \State $y_x \gets$ corresponding variable of $x$ in $simpModel$ 
        \If {$y_x$ is a fixed variable}
            \State $varMap[x] \gets$ the fixed value of $y_x$\label{algo:fixedVar}
        \ElsIf {$y_x$ is a free variable}
            \State $varMap[x] \gets$ new Boolean variable in $simpInst$\label{createNewVar}
        \ElsIf {$y_x$ is a simple aggregated variable} 
            \State $y_z \gets$  final variable in the aggregation chain\label{algo:simpAgg1}
            \State create $varMap[z]$ if it was not created
            \State $varMap[x] \gets$ $varMap(z)$ or $\neg varMap(z)$ according to the aggregation relationship\label{algo:simpAgg2}
        \ElsIf {$y_x$ is a multiple aggregated variable}
            \State $varMap[x] \gets$ new Boolean variable in $simpInst$\label{newVarForAgg}
            \State Encode the aggregation constraint with a pseudo-Boolean encoding\label{pbForAgg}
        \EndIf
    \EndFor
  \end{algorithmic}
\end{algorithm}

Here is an example of processing simple aggregation variables: consider three variables in $simpModel$ with the aggregation relationships $(y_1 = 1-y_2)$ and $(y_2 = y_3)$. In this case, only one new Boolean variable $v_1$ is created in $simpInst$ to represent $y_1$, $y_2$ and $y_3$, by implementing the mapping $\{y_1 \rightarrow \neg v_1, y_2 \rightarrow v_1, y_3 \rightarrow v_1\}$ when transforming $simpModel$ to $simpInst$, which preserves $(y_1 = 1 - y_2)$ and $(y_2 = y_3)$. Together with variable fixing, this operation significantly reduces the number of variables in $simpInst$ w.r.t. $originInst$, as will be showcased empirically in Section~\ref{Exp}.

\subsection{Constraint Encoding}

We use the SCIP solver \cite{achterberg_scip_2009} to preprocess $originModel$ as it is an open-source mixed-integer programming solver broadly used in MaxSAT evaluations. The obtained $simpModel$ usually contains various types of constraints, as listed in Table~\ref{tab:constraints_encoding}. \textit{Logical OR} and \textit{Logical AND} constraints are directly encoded into CNF. \textit{Setppc} and \textit{Linear} constraints are encoded into CNF using the methods for at-most-one and pseudo-Boolean constraints in the PBLib library \cite{PBLib_Springer_2015}, respectively. We use the default configuration in PBLib, allowing it to automatically select the most suitable encoding (such as Binary Decision Diagrams (BDD)~\cite{BDD_encoding} or Adder Networks~\cite{PBEncoding_Niklas}, among others) based on the properties of the constraints. 

There are also constraints specifically designed for SCIP, which are not listed in Table~\ref{tab:constraints_encoding}. In our experiment, we found that about 8\% of instances contain the \textit{orbitope} constraint, which arises from orbitopal fixing~\cite{Orbitopal_fixing_Volker}. Given the complexity of encoding the orbitope constraint in CNF, instances containing it will not be encoded into $simpInst$, and the $originInst$ will be solved instead. 
The objective function of $simpModel$ is defined in the equation $f^{'}_{(S)} = Maximize \sum_{c \in S} w_c^{'} \cdot z_c^{'}$, where $S$ is the set of soft clauses,  $z_c^{'}$ is the decision variable in $simpModel$, and $w_c^{'}$ is the corresponding coefficient. We encode $f^{'}_{(S)}$ into soft clauses using the following method: if a coefficient $w_c^{'}$ of a decision variable $z_c^{'}$ is positive, then $z_c^{'}$ is added as a soft clause with weight $w_c^{'}$, otherwise, $\neg z_c^{'}$ is added with weight $-w_c^{'}$.

\section{Experimental Results} \label{Exp}

\subsection{Experimental Protocol}

We use state-of-the-art ILP and MaxSAT solvers for our experiments. More specifically, SCIP~(version 9.1.1)~\cite{achterberg_scip_2009} is used as the ILP solver for preprocessing. For MaxSAT, we select the top-performing solvers from the MaxSAT evaluation 2024. In the unweighted category, the leading solvers are WMaxCDCL-OpenWbo1200 \cite{WMaxCDCL_2024_Li}, MaxCDCL-OpenWbo300~\cite{MaxCDCL_2024_Li}, and UWrMaxSat-SCIP-MaxPre~\cite{UWrMaxSAT_2024_Marek}, which are ranked as the top three. In the weighted category, these solvers are CASHWMaxSAT-DisjCom-S6~\cite{CashMaxSAT_2024_Pan}, UWrMaxSat-SCIP~\cite{UWrMaxSAT_2024_Marek}, and EvalMaxSAT~\cite{EvalMaxSAT_2024_Florent}. WMaxCDCL-OpenWbo1200 (MaxCDCL-OpenWbo300) runs OpenWbo~\cite{OpenWBO_Sinz} for 1200s (300s), followed by WMaxCDCL (MaxCDCL) for 2400s (3300s) to solve an instance; CASHWMaxSAT-DisjCom-S6 and EvalMaxSAT first run SCIP for 600s (400s), followed by CASHWMaxSAT (EvalMaxSAT); UWrMaxSat-SCIP uses a more complex portfolio strategy: it first runs UWrMaxSat for several seconds to get upper and lower bounds, then calls SCIP with the computed bounds and a time limit; if SCIP fails to produce the optimal solution, it runs UWrMaxSat again with the new bounds from SCIP for the rest of the time.

The benchmark MaxSAT instances are sourced from the unweighted and weighted categories of MaxSAT evaluations from 2019 to 2024 (MS19-MS24 and WMS19-WMS24, respectively). To avoid counting the duplicated instances twice, we removed from (W)MS$k$ for $k>19$ the instances also occurring in previous years from 2019. The 2209 unweighted (2223 weighted) partial MaxSAT instances are from 74 (63) families. Each family represents a specific optimization problem encoded into MaxSAT, from different fields related to combinatorial optimization and AI, making the tests and our observations more comprehensive and credible.

The computations are performed on the MatriCS platform\footnote{https://www.matrics.u-picardie.fr/} equipped with AMD EPYC 7502 Processors (2.5 GHz) and the Linux system. Each solver is given one CPU and 31 GB of RAM, with a time limit of 3600 seconds, following the MaxSAT Evaluation setting for time.

\begin{figure}[!t]
  \centering
  \begin{subfigure}[b]{0.5\textwidth}
    \centering
    \includegraphics[width=\textwidth]{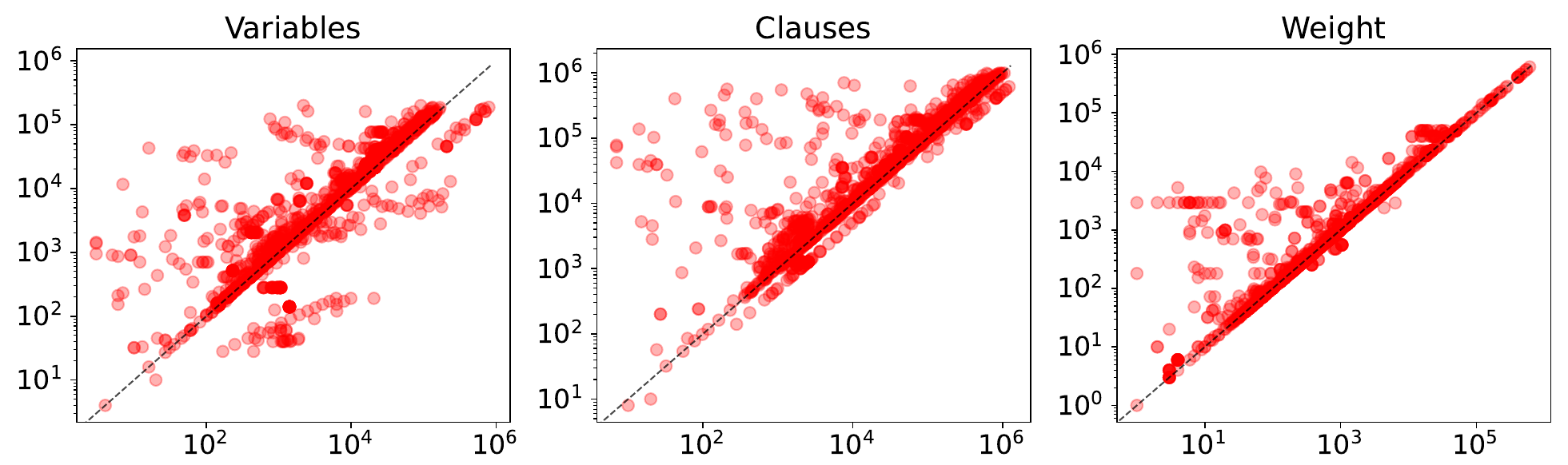}
    \caption{\centering Unweighted category (MS19-MS24)}
  \end{subfigure}
  \vskip\baselineskip
  \begin{subfigure}[b]{0.5\textwidth}
    \centering
    \includegraphics[width=\textwidth]{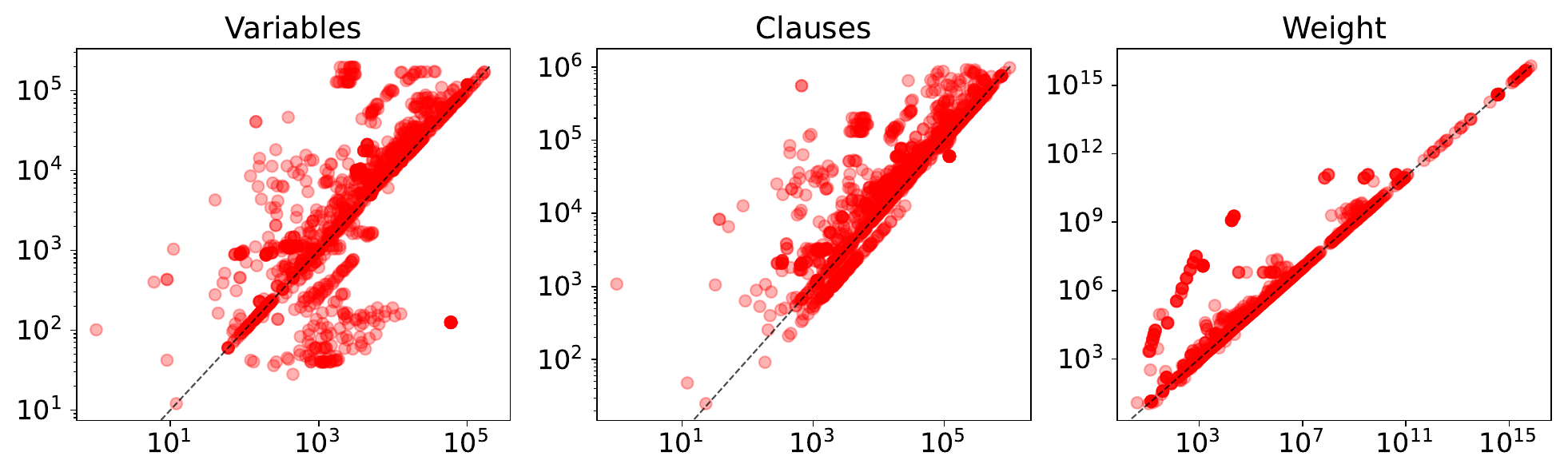}
    \caption{\centering Weighted category (WMS19-WMS24)}
  \end{subfigure}
  \caption{Comparison between $originInst$ (vertical axes) and $simpInst$ (horizontal axes) in terms of variables, clauses, and the sum of soft weights. The axes are on a logarithmic scale.}
  \label{fig:ILP_preprocessing_scatter}
\end{figure}

\subsection{Impact of ILP Preprocessing} \label{ILP_statistic}

We first evaluated the impact of ILP preprocessing on the size of simplified instances. Figure~\ref{fig:ILP_preprocessing_scatter} compares the number of variables and clauses (including hard and soft clauses), and the total weight of soft clauses between $originInst$ and $simpInst$. In each subgraph, the vertical axis represents $originInst$ and the horizontal axis represents $simpInst$. The diagonal line serves as a reference, where a point above the line signifies that $originInst$ is larger than the corresponding $simpInst$ in terms of variables, clauses, or weights. These subgraphs demonstrate that ILP preprocessing can help reduce the variables and clauses for most of the instances, and more than half of the instances get a reduction of the total soft weights. More specifically, in the unweighted category, 72.66\%, 74.95\%, and 54.03\% of simplified instances result in fewer variables, clauses, and weight, respectively. In the weighted category, the corresponding percentages are 73.78\%, 72.84\%, and 61.03\%. However, the encoding of the at-most-one and the pseudo-Boolean constraints might introduce new variables and clauses in $simpInst$, making some simplified instances larger than the corresponding original instances. Specifically, 13.18\% and 13.64\% $simpInst$ have a larger number of variables and clauses, respectively, whereas the corresponding percentages are 14.89\% and 18.58\% for the weighted category. Since an increase in the number of variables and clauses is usually associated with a decline in MaxSAT solving efficiency, only the smaller instances are selected to be solved by the MaxSAT solver; otherwise, the original instances are used (see Section~\ref{ThreeStagesMethod}).

\begin{table}[!t]
    \centering
    \caption{Statistics of two groups of instances w.r.t. the SCIP preprocessing. $\Delta vars$ ($\Delta clauses$) represents the change in the number of variables (clauses) from $originInst$ to $simpInst$. $simpleAggrRatio$ is the ratio of simple aggregated variables to all the aggregated variables. Metrics are calculated by the mean value of instances in each group.}
    \label{tab:simplified_metrics}
    \begin{tabular}{ccccc}
    \toprule
    \multicolumn{1}{l}{}     & \multicolumn{2}{c}{Unweighted category} & \multicolumn{2}{c}{Weighted category} \\
    \cmidrule(lr){2-3} \cmidrule(lr){4-5}
    Statistic metrics     & Smaller     & Bigger    & Smaller     & Bigger     \\
    \midrule
    \#Instances            & 1085        & 436      & 980         & 508        \\
    $\Delta vars (\%)$     & -22.61\%    & +116.27\% & -25.04\%   & +666.71\%   \\
    $\Delta clauses (\%)$  & -21.83\%   & +7.35\%   & -25.64\%   & +8.83\% \\
    $fixedVarsRate$     & 18.66\%     & 3.64\%      & 19.61\%     & 16.30\%    \\
    $aggrVarsRate$ & 26.92\%     & 21.52\%    & 27.68\%     & 18.36\%    \\
    $simpleAggrRatio$   & 99.49\%  & 79.54\%   & 99.22\%      & 96.64\%  \\
    $preprocessingTime$  & 15.36s      & 30.0s      & 14.26s      & 19.65s     \\
    \bottomrule
    \end{tabular}
\end{table}

Table~\ref{tab:simplified_metrics} shows the metrics for ILP preprocessing, where instances are partitioned into the "Smaller" and "Bigger" groups: if $simpInst$ contains fewer variables and fewer hard clauses than $originInst$, the instance is in the "Smaller" group; otherwise, it is in the "Bigger" group. There are some instances not listed in the table, because these instances are either too large (with more than 200,000 variables or 1,000,000 clauses) to be preprocessed or cannot be encoded into $simpInst$. The results indicate that ILP preprocessing causes significant changes in the number of variables ($\Delta vars$) and clauses ($\Delta clauses$). Notably, the number of variables decreased by 22.61\% (25.04\%) for unweighted (weighted) instances in the "Smaller" group, whereas the "Bigger" group showed an increase of 116.27\% (666.71\%). Furthermore, for the unweighted instances, the fixed variables rate ($fixedVarsRate$) is substantially higher in the "Smaller" group compared to the "Bigger" group (18.66\% vs. 3.64\%). Similarly, for the weighted instances, the aggregated variables rate ($aggrVarsRate$) is higher in the "Smaller" group (27.68\% vs. 18.36\%). This demonstrates that ILP preprocessing techniques, through variable fixing and aggregation, effectively reduce the size of $simpInst$. Additionally, within the "Smaller" group, the ratio of simple aggregated variables ($simpleAggrRatio$) is larger than 99\%. This indicates that the vast majority of these aggregated variables belong to the forms: $y_x = y_1$ or $y_x = 1-y_1$, either mirroring a variable or its negation. Our variable encoding (see lines \ref{algo:simpAgg1}-\ref{algo:simpAgg2} in Algorithm~\ref{alg:encode_variables}) leverages this property to eliminate at least half of the simple aggregated variables, thereby reducing the variables for instances in the "Smaller" group. Finally, SCIP preprocessing time ($preprocessingTime$) is negligible compared to the total time limit (less than 1\% out of 3600s).

\subsection{Performance of MaxSAT solvers}

\begin{table*}[!t]
    \centering
    \caption{Number of instances solved by each solver within 3600s. Values before the parentheses indicate the result of baseline solvers, and values in the parentheses indicate the additional solved instances after applying ILP preprocessing (+simp) to the baseline solvers. RAR2 score (smaller is better) is the average CPU time for all the tested instances, and the punishment for each unsolved instance is twice the testing time (7200s).}
    \label{tab:solved_instances_overview}
    \begin{tabular}{ccccccccc}
    \toprule
    Unweighted category   & MS19                 & MS20                 & MS21                 & MS22                 & MS23                   & MS24                 & \textbf{Total}    & \textbf{PAR2}     \\
    \midrule
    \#Instances           & 599                  & 401                  & 448                  & 254                  & 260                  & 247                  & \textbf{2209}     & -   \\
    SCIP                  & 235                  & 203                  & 208                  & 102                  & 118                  & 88                   & \textbf{954}      & 4222.46   \\
    WMaxCDCL-OpenWbo1200(+simp)   & 446(+10)              & 306(+1)              & 339(+2)              & 183(+1)              & 177(+1)              & 199(0)      & \textbf{1650(+15)} & 2036.38(-2.34\%)   \\
    MaxCDCL-OpenWbo300(+simp)     & 441(+10)              & 315(-1)              & 344(+4)              & 186(+1)              & 177(+1)              & 197(+1)     & \textbf{1660(+16)} & 1939.24(-2.14\%)  \\
    UWrMaxSat-SCIP-Maxpre(+simp)  & 445(0)              & 329(0)               & 352(0)               & 183(0)               & 188(+2)               & 175(+1)      & \textbf{1672(+3)}  & 1857.28(-0.13\%)  \\
    \midrule   
    \midrule
    Weighted  category      & WMS19                & WMS20                & WMS21                & WMS22                & WMS23                & WMS24                & \textbf{Total}   & \textbf{PAR2}      \\
    \midrule
    \#Instances              & 586                  & 433                  & 491                  & 291                  & 218                  & 204                  & \textbf{2223}      & -    \\
    SCIP                         & 228                  & 239                  & 238                  & 122                  & 117                  & 101                  & \textbf{1045}   &  3875.32    \\
    CASHWMaxSAT-DisjCom-S6(+simp) & 410(+4)              & 349(-1)              & 405(-2)              & 214(0)              & 167(+1)               & 151(0)              & \textbf{1696(+2)}   & 1907.64(-0.41\%)  \\
   UWrMaxSat-SCIP(+simp)         & 404(-3)              & 347(0)               & 398(0)               & 209(+1)              & 163(0)               & 146(0)               & \textbf{1667(-2)}   & 1841.73(+0.34\%)   \\
    EvalMaxSAT(+simp)             & 390(+3)              & 338(+1)              & 396(-2)              & 212(+1)              & 162(+1)               & 148(+1)               & \textbf{1646(+5)} & 2031.76(-0.08\%)  \\
    WMaxCDCL-OpenWbo1200(+simp)   & 391(+1)              & 337(+4)              & 397(+2)              & 210(-1)              & 154(0)               & 147(0)               & \textbf{1636(+6)}  & 2119.2(-1.38\%) \\
    \bottomrule
    \end{tabular}
\end{table*}

\begin{figure}[!b]
  \centering
  \begin{subfigure}[b]{0.5\textwidth}
    \centering
    \includegraphics[width=0.9\textwidth]{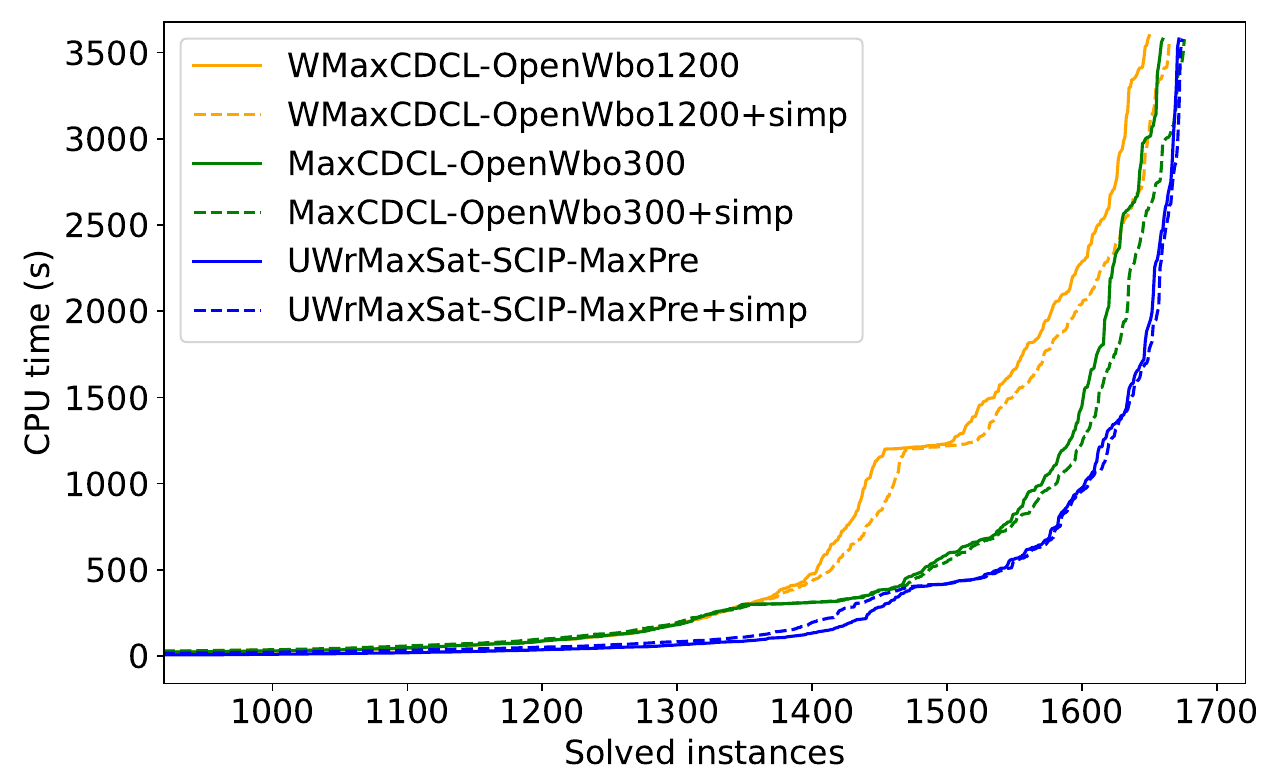}
    \caption{\centering Unweighted category (MS19-MS24)}
  \end{subfigure}
  \vskip\baselineskip
  \begin{subfigure}[b]{0.5\textwidth}
    \centering
    \includegraphics[width=0.9\textwidth]{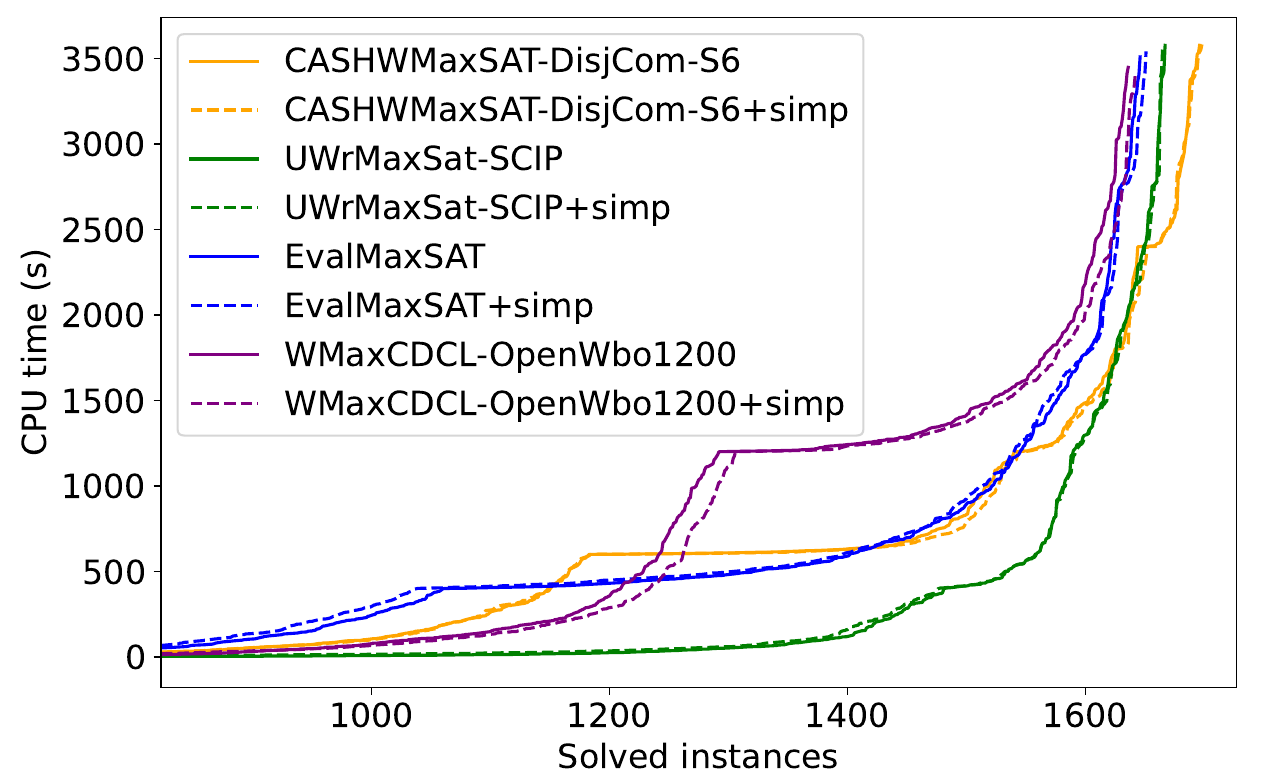}
    \caption{\centering Weighted category (WMS19-WMS24)}
  \end{subfigure}
  \caption{Number of solved instances changes with solving time (in seconds). Solid lines in the subfigures represent the baseline solvers, and the dashed lines represent the baseline solvers with ILP preprocessing (+simp).}
  \label{fig:solved_instances_cpu_time}
\end{figure}

Table \ref{tab:solved_instances_overview} compares the number of solved instances by each solver with the ILP preprocessing (+simp) and without. Figure~\ref{fig:solved_instances_cpu_time} presents the number of solved instances along with their CPU time costs for each solver. For the unweighted category, ILP preprocessing helps to improve the efficiency of all the top-performing solvers. Moreover, our method enables WMaxCDCL-OpenWbo1200, the winner of the MaxSAT evaluation 2024, to solve 15 additional instances. For the weighted category, 3 out of 4 solvers managed to solve more instances with ILP preprocessing. The only exception, with two fewer instances, is UWrMaxSat-SCIP, which uses a very complex portfolio strategy, and ILP preprocessing may conflict with some of its heuristic settings.
We clarify that these improvements are significant because MaxSAT solving has reached a high level of maturity, making further improvements increasingly hard. As a matter of fact, in the unweighted category of the MaxSAT evaluation 2024, the winner WMaxCDCL-OpenWbo1200 solves only 2 (+0.48\%) instances more than the 2nd ranked solver MaxCDCL-OpenWbo300, which solves in turn 4 (+0.96\%) instances more than the 3rd ranked solver UWrMaxSat-SCIP-Maxpre, out of a total of 553 instances\footnote{https://maxsat-evaluations.github.io/2024/rankings.html}, while our method helps WMaxCDCL-OpenWbo1200 (MaxCDCL-OpenWbo300) to solve +0.91\% (+0.96\%) more unweighted partial MaxSAT instances.

Among the baseline MaxSAT solvers in our experiment, WMaxCDCL-OpenWbo1200 and MaxCDCL-OpenWbo300 are the solvers that do not use SCIP in their portfolios. To see if SCIP preprocessing is compatible with the SCIP portfolio, we tested four new solvers: WMO+simp+S4 (WMO+simp+S1) runs SCIP as an entire solver for 400s (100s), then WMaxCDCL-OpenWbo1200 with the SCIP preprocessing for 3200s (3500s); MO+simp+S4 and MO+simp+S1 are similar but use MaxCDCL instead of WMaxCDCL. The test results are represented in Table \ref{tab:Portfolio_with_SCIP}, indicating that the versions running SCIP for 100s give very good results and allow MO+simp and WMO+simp to solve 12 and 14 (3 in the weighted category) more instances, respectively. But the versions running SCIP for 400s give worse results. This indicates that SCIP preprocessing does not fully bridge the gap of using SCIP as a portfolio solver, but it greatly reduces the need to call it as such. Additionally, the PAR2 score of (W)MO+simp+S4 is much worse than (W)MO+simp and (W)MO+simp+S1. For example, the PAR2 score for MO+simp+S4 increased by 8.23\% compared to MO+simp. Recall that running SCIP 400s as an entire solver in a portfolio means that we generally lose 400s for most instances for which SCIP is not effective, but we just lose 100s if we run SCIP only 100s in a portfolio.

\begin{table*}[!t]
  \centering
  \caption{Number of instances solved by WMaxCDCL-OpenWbo1200 (WMO+simp) or MaxCDCL-OpenWbo300 (MO+simp) with the SCIP solver as a portfolio. S1 (S4) corresponds to running SCIP for 100 (400) seconds in the portfolio.}
  \label{tab:Portfolio_with_SCIP}

    \begin{tabular}{ccccccccc}
    
    \toprule
    Unweighted category            &MS19    & MS20    & MS21    & MS22    & MS23    & MS24    & \textbf{Total}  & \textbf{PAR2}     \\
    \midrule
    WMO+simp(+S4/+S1) & 456(+4/+6) & 307(+2/+3) & 341(-2/+2) & 184(-1/0) & 178(+1/0)  & 199(+1/+1)  & \textbf{1665(+5/+12)}  & 1988.74(+6.38\%/-0.32\%)\\
    MO+simp(+S4/+S1)   & 451(+6/+8) & 314(+3/+4) & 348(-4/0) & 187(-2/-1)  & 178(+2/+2) & 198(+2/+1) & \textbf{1676(+7/+14)}  & 1897.77(+8.23\%/+0.36\%) \\
    \midrule
    \midrule
    Weighted category            &WMS19    & WMS20    & WMS21    & WMS22    & WMS23    & WMS24    & \textbf{Total}  & \textbf{PAR2}      \\
    \midrule
    WMO+simp(+S4/+S1) & 392(+2/+1) & 341(-2/+1) & 399(0/0) & 209(+1/+1) & 154(0/0) & 147(-1/0) & \textbf{1642(0/+3)} & 2090.03(+4.72\%/-2.74\%) \\
    \bottomrule
    \end{tabular}
\end{table*}

\section{Conclusion} \label{Conc}

In this paper, we proposed a methodology to integrate ILP preprocessing techniques into the MaxSAT solving pipeline, and investigated its impact on the top-performing MaxSAT solvers. The experimental results show that ILP preprocessing techniques help to simplify the MaxSAT instances by reducing the number of variables and clauses; they also enable 5 out of 6 state-of-the-art MaxSAT solvers to solve more instances within a time limit of 3600 seconds. Moreover, our approach enables WMaxCDCL-OpenWbo1200, the winner of the MaxSAT evaluation 2024 in the unweighted track, to solve 15 (+0.91\%) additional instances on this track. Our results also indicate that the ILP preprocessing reduces the need to call the ILP solver in a portfolio setting within MaxSAT solvers, including WMaxCDCL-OpenWbo1200 and MaxCDCL-OpenWbo300, since running SCIP for a shorter time achieves better results. 

Future work encompasses two primary directions. First, we will enhance the encoding of the ILP model into a MaxSAT instance by exploring more advanced encoding methods, potentially further reducing the size of the simplified instances. Second, we plan to evaluate ILP preprocessing techniques individually to identify those that enable us to significantly improve MaxSAT solving. Subsequently, we aim to directly integrate the most effective techniques into a MaxSAT solver to improve preprocessing efficiency.

\section*{Acknowlegement}
This work is partially supported by the ANR-19-CHIA0013-01 chair (MASSAL’IA), co-funded by the French national research agency and the French electricity distribution network operator Enedis, and by the ANR-24-CE23-6126 project (BforSAT).
It was also granted access to the HPC resources of “Plateforme MatriCS” within the University of Picardie Jules Verne, which is co-financed by the European Union with the European Regional Development Fund (FEDER) and the Hauts-De-France Regional Council, among others.


\bibliography{references_short}

@incollection{li:hal-04321437,
  title       = {{Chapter 23. MaxSAT, Hard and Soft Constraints}},
  author      = {Li, Chu Min and Many{\`a}, Felip},
  booktitle   = {{Handbook of Satisfiability}},
  publisher   = {{IOS Press}},
  series      = {Frontiers in Artificial Intelligence and Applications},
  year        = {2021},
  month       = Feb,
  hal_id      = {hal-04321437},
  hal_version = {v1}
}

@article{li_combining_2021,
  title        = {Combining {Clause} {Learning} and {Branch} and {Bound} for {MaxSAT}},
  volume       = {210},
  issn         = {1868-8969},
  language     = {en},
  urldate      = {2025-01-26},
  journal      = {LIPIcs, Volume 210, CP 2021},
  author       = {Li, Chu-Min and Xu, Zhenxing and Coll, Jordi and Manyà, Felip and Habet, Djamal and He, Kun},
  collaborator = {Michel, Laurent D.},
  year         = {2021},
  pages        = {38:1--38:18}
}

@inproceedings{minsat,
  title     = {Exact Minsat Solving},
  author    = {Li, Chu-Min and Manya, Felip and Quan, Zhe and Zhu, Zhu},
  booktitle = {SAT-2010},
  year      = {2010},
  pages     = {363-368}
}

@article{AIJChumin,
    author = {Chu-Min Li and Fan Xiao and Mao Luo and Felip Many\`a and Zhipeng L\"u and Yu Li},
    title = {{Clause vivification by unit propagation in CDCL SAT solvers}},
    journal = {Artificial Intelligence},
    volume = {279},
    pages = {103197},
    year = {2020}
}

@inproceedings{Anstegui2013SolvingP,
  title     = {Solving (Weighted) Partial MaxSAT with {ILP}},
  author    = {Carlos Ans{\'o}tegui and Joel Gab{\`a}s},
  booktitle = {Integration of AI and OR Techniques in Constraint Programming},
  year      = {2013},
}

@article{achterberg_scip_2009,
  title   = {{SCIP}: solving constraint integer programs},
  volume  = {1},
  issn    = {1867-2957},
  number  = {1},
  journal = {Mathematical Programming Computation},
  author  = {Achterberg, Tobias},
  month   = jul,
  year    = {2009},
  pages   = {1--41}
}

@article{savelsbergh_preprocessing_1994,
  title    = {Preprocessing and {Probing} {Techniques} for {Mixed} {Integer} {Programming} {Problems}},
  volume   = {6},
  issn     = {0899-1499, 2326-3245},
  language = {en},
  number   = {4},
  urldate  = {2025-02-23},
  journal  = {ORSA Journal on Computing},
  author   = {Savelsbergh, M. W. P.},
  month    = nov,
  year     = {1994},
  pages    = {445--454}
}

@book{maxsat_evaluation_2024,
  title     = {MaxSAT Evaluation 2024: Solver and Benchmark Descriptions},
  editor    = {Jeremias Berg and Matti J{\"a}rvisalo and Ruben Martins and Andreas Niskanen and Tobias Paxian},
  year      = {2024},
  language  = {English},
  publisher = {University of Helsinki},
  address   = {Finland}
}

@inproceedings{DBLP:conf/cp/CherifSLLD24,
  author       = {Sami Cherif and
                  Heythem Sattoutah and
                  Chu{-}Min Li and
                  Corinne Lucet and
                  Laure Brisoux Devendeville},
  title        = {Minimizing Working-Group Conflicts in Conference Session Scheduling
                  Through Maximum Satisfiability},
  booktitle    = {{CP} 2024, September 2-6, 2024, Girona, Spain},
  bibsource    = {dblp computer science bibliography, https://dblp.org}
}

@inproceedings{DBLP:conf/ictai/ZhengCS24,
  author       = {Zhifei Zheng and
                  Sami Cherif and
                  Rui Sa Shibasaki},
  title        = {Optimizing Power Peaks in Simple Assembly Line Balancing Through Maximum
                  Satisfiability},
  booktitle    = {{ICTAI} 2024, Herndon, VA, USA, October 28-30, 2024},
  pages        = {363--370},
  publisher    = {{IEEE}},
  year         = {2024},
  bibsource    = {dblp computer science bibliography, https://dblp.org}
}

@INPROCEEDINGS{SOCdesignApplication,
  author={Safarpour, Sean and Mangassarian, Hratch and Veneris, Andreas and Liffiton, Mark H. and Sakallah, Karem A.},
  booktitle={FMCAD'07}, 
  title={Improved Design Debugging Using Maximum Satisfiability}, 
  year={2007},
  volume={},
  number={},
  pages={13-19},
}

@article{MaxSATforTimetabling,
title = {MaxSAT-based large neighborhood search for high school timetabling},
journal = {Computers \& Operations Research},
volume = {78},
pages = {172-180},
year = {2017},
issn = {0305-0548},
author = {Emir Demirović and Nysret Musliu}
}

@InProceedings{PBLib_Springer_2015,
    author="Philipp, Tobias
    and Steinke, Peter",
    editor="Heule, Marijn
    and Weaver, Sean",
    title="PBLib -- A Library for Encoding Pseudo-Boolean Constraints into CNF",
    booktitle="SAT 2015",
    year="2015",
    publisher="Springer International Publishing",
    address="Cham",
    pages="9--16",
    isbn="978-3-319-24318-4"
}

@inbook{handbookofSAT_chapter24_Fahiem,
    title = "Maximum Satisfiability",
    author = "Fahiem Bacchus and Matti J{\"a}rvisalo and Ruben Martins",
    year = "2021",
    language = "English",
    isbn = "978-1-64368-160-3",
    series = "Frontiers in Artificial Intelligence and Applications",
    publisher = "IOS PRESS",
    pages = "929 -- 991",
    editor = "Armin Biere and Marijn Heule and {van Maaren}, Hans and Toby Walsh",
    booktitle = "Handbook of Satisfiability",
    address = "Netherlands",
    edition = "2",
}

@inbook{SAT_preprocessing_Armin,
    title = "Preprocessing in SAT Solving",
    author = "Armin Biere and Matti J{\"a}rvisalo and Benjamin Kiesl",
    year = "2021",
    language = "English",
    isbn = "978-1-64368-160-3",
    series = "Frontiers in Artificial Intelligence and Applications",
    publisher = "IOS PRESS",
    pages = "391 -- 435",
    editor = "Armin Biere and Marijn Heule and {van Maaren}, Hans and Toby Walsh",
    booktitle = "Handbook of Satisfiability",
    address = "Netherlands",
    edition = "2",
}

@InProceedings{MaxSAT_preprocessing_Argelich,
    author="Argelich, Josep
    and Li, Chu Min
    and Many{\`a}, Felip",
    editor="Kleine B{\"u}ning, Hans
    and Zhao, Xishun",
    title="A Preprocessor for Max-SAT Solvers",
    booktitle="SAT 2008",
    year="2008",
    publisher="Springer Berlin Heidelberg",
    address="Berlin, Heidelberg",
    pages="15--20",
    isbn="978-3-540-79719-7"
}

@article{ILP_presolve_Achterberg,
    author = {Achterberg, Tobias and Bixby, Robert and Gu, Zonghao and Rothberg, Edward and Weninger, Dieter},
    year = {2019},
    month = {11},
    pages = {},
    title = {Presolve Reductions in Mixed Integer Programming},
    volume = {32},
    journal = {INFORMS Journal on Computing},
}

@article{MaxCDCL_2024_Li,
  title={MaxCDCL in MaxSAT Evaluation 2024},
  author={Li, Chu Min and Li, Shuolin and Coll, Jordi and Habet, Djamal and Manyà, Felip and He, Kun},
  journal={MaxSAT Evaluation 2024 Solver and Benchmark Descriptions},
  pages={15-16},
  year={2024},
  publisher = "Department of Computer Science, University of Helsinki",
  address = "Finland",
}

@article{WMaxCDCL_2024_Li,
  title={WMaxCDCL in MaxSAT Evaluation 2024},
  author={Li, Shuolin and Li, Chu Min and Coll, Jordi and Habet, Djamal and Manyà, Felip and He, Kun},
  journal={MaxSAT Evaluation 2024 Solver and Benchmark Descriptions},
  pages={17-18},
  year={2024},
  publisher = "Department of Computer Science, University of Helsinki",
  address = "Finland",
}

@article{UWrMaxSAT_2024_Marek,
  title={UWrMaxSat Entering the MaxSAT Evaluation 2024},
  author={Piotrów, Marek},
  journal={MaxSAT Evaluation 2024 Solver and Benchmark Descriptions},
  pages={27-28},
  year={2024},
  publisher = "Department of Computer Science, University of Helsinki",
  address = "Finland",
}

@article{EvalMaxSAT_2024_Florent,
  title={EvalMaxSAT 2024},
  author={Avellaneda, Florent},
  journal={MaxSAT Evaluation 2024 Solver and Benchmark Descriptions},
  pages={8},
  year={2024},
  series = "Department of Computer Science Series of Publications B",
  publisher = "Department of Computer Science, University of Helsinki",
  address = "Finland",
}

@article{CashMaxSAT_2024_Pan,
  title={CASHWMaxSAT-DisjCad: Solver Description},
  author={Pan, Shiwei and Wang, Yiyuan and Cai, Shaowei and Li, Jiangnan and Zhu, Wenbo and Yin, Minghao},
  journal={MaxSAT Evaluation 2024 Solver and Benchmark Descriptions},
  pages={25},
  year={2024},
  series = "Department of Computer Science Series of Publications B",
  publisher = "Department of Computer Science, University of Helsinki",
  address = "Finland",
}

@inproceedings{maxsat_for_ai_hao,
  TITLE = {{Optimizing Binary Decision Diagrams with MaxSAT for Classification}},
  AUTHOR = {Hu, Hao and Huguet, Marie-Jos{\'e} and Siala, Mohamed},
  BOOKTITLE = {{AAAI 2022}},
  HAL_LOCAL_REFERENCE = {Rapport LAAS n{\textdegree} 22158},
  YEAR = {2022},
  MONTH = Feb,
  HAL_ID = {hal-03667549},
  HAL_VERSION = {v1},
}

@article{Chen2010AutomatedDD,
  title={Automated Design Debugging With Maximum Satisfiability},
  author={{Yibin Chen and Sean Safarpour and Joao Marques-Silva and Andreas G. Veneris}},
  journal={IEEE Transactions on Computer-Aided Design of Integrated Circuits and Systems},
  year={2010},
  volume={29},
  pages={1804-1817},
}

@InProceedings{OpenWBO_Sinz,
    author="Martins, Ruben
    and Manquinho, Vasco
    and Lynce, In{\^e}s",
    editor="Sinz, Carsten
    and Egly, Uwe",
    title="Open-WBO: A Modular MaxSAT Solver,",
    booktitle="SAT 2014",
    year="2014",
    publisher="Springer International Publishing",
    address="Cham",
    pages="438--445",
}

@article{MaxSAT_AI_Alexey,
  title={XAI-MinDSet2: Explainable AI with MaxSAT},
  author={Alexey Ignatiev and Joao Marques-Silva},
  journal={MaxSAT Evaluation 2018: Solver and Benchmark Descriptions},
  pages={43},
  year={2018},
  series = "Department of Computer Science Series of Publications B",
  publisher = "Department of Computer Science, University of Helsinki",
  address = "Finland",
}

@article{Orbitopal_fixing_Volker,
title = {Orbitopal fixing},
journal = {Discrete Optimization},
volume = {8},
number = {4},
pages = {595-610},
year = {2011},
issn = {1572-5286},
author = {Volker Kaibel and Matthias Peinhardt and Marc E. Pfetsch},
}

@InProceedings{BDD_encoding,
author="Ab{\'i}o, Ignasi
and Nieuwenhuis, Robert
and Oliveras, Albert
and Rodr{\'i}guez-Carbonell, Enric",
editor="Sakallah, Karem A.and Simon, Laurent",
title="BDDs for Pseudo-Boolean Constraints -- Revisited",
booktitle="SAT 2011",
year="2011",
publisher="Springer Berlin Heidelberg",
address="Berlin, Heidelberg",
pages="61--75",
}

@article{PBEncoding_Niklas,
author = {Niklas Eén and Niklas Sörensson},
title ={Translating Pseudo-Boolean Constraints into SAT},
journal = {Journal on Satisfiability, Boolean Modelling and Computation},
volume = {2},
number = {1-4},
pages = {1-26},
year = {2006},
}

@incollection{gaspers_maxpre_2017,
	address = {Cham},
	title = {{MaxPre}: {An} {Extended} {MaxSAT} {Preprocessor}},
	volume = {10491},
	copyright = {http://www.springer.com/tdm},
	isbn = {978-3-319-66262-6 978-3-319-66263-3},
	shorttitle = {{MaxPre}},
	language = {en},
	urldate = {2025-03-03},
	booktitle = {{SAT} 2017},
	publisher = {Springer International Publishing},
	author = {Korhonen, Tuukka and Berg, Jeremias and Saikko, Paul and Järvisalo, Matti},
	editor = {Gaspers, Serge and Walsh, Toby},
	year = {2017},
	pages = {449--456},
}

@article{ignatiev_rc2_2019,
	title = {{RC2}: an {Efficient} {MaxSAT} {Solver}},
	volume = {11},
	copyright = {https://creativecommons.org/licenses/by-nc/4.0/},
	issn = {15740617},
	shorttitle = {{RC2}},
	language = {en},
	number = {1},
	urldate = {2025-01-26},
	journal = {Journal on Satisfiability, Boolean Modeling and Computation},
	author = {Ignatiev, Alexey and Morgado, Antonio and Marques-Silva, Joao},
	editor = {Seidl, Martina and Pulina, Luca},
	month = sep,
	year = {2019},
	pages = {53--64},
}

@InProceedings{Coprocessor2,
    author="Manthey, Norbert",
    editor="Cimatti, Alessandro
    and Sebastiani, Roberto",
    title="Coprocessor 2.0 -- A Flexible CNF Simplifier",
    booktitle="Theory and Applications of Satisfiability Testing -- SAT 2012",
    year="2012",
    publisher="Springer Berlin Heidelberg",
    address="Berlin, Heidelberg",
    pages="436--441",
}

@article{wmaxcdcl_2025_jordi,
  title={Solving weighted Maximum Satisfiability with Branch and Bound and clause learning},
  author={Coll, Jordi and Li, Chu-Min and Li, Shuolin and Habet, Djamal and Many{\`a}, Felip},
  journal={Computers \& Operations Research},
  pages={107195},
  year={2025},
  publisher={Elsevier}
}

@inproceedings{wmaxcdcl_2025_li,
  title={Improving the lower bound in branch-and-bound algorithms for maxsat},
  author={Li, Shuolin and Li, Chu-Min and Coll, Jordi and Habet, Djamal and Many{\`a}, Felip},
  booktitle={AAAI 2025},
  volume={39},
  number={11},
  year={2025}
}

@inproceedings{esa_2025_li,
  title={A new variable ordering for in-processing bounded variable elimination in SAT solvers},
  author={Li, Shuolin and Li, Chu-Min and Luo, Mao and Coll, Jordi and Habet, Djamal and Many{\`a}, Felip},
  booktitle={IJCAI 2023},
  year={2023}
}

@inproceedings{vivification_2017_luo,
  title={An effective learnt clause minimization approach for CDCL SAT solvers},
  author={Luo, Mao and Li, Chu-Min and Xiao, Fan and Manya, Felip and L{\"u}, Zhipeng},
  booktitle={IJCAI 17},
  pages={703--711},
  year={2017}
}

@inproceedings{tableauMaxSAT2016,
  title={A Clause Tableau Calculus for MaxSAT},
  author={Li, Chu-Min and Manyà, Felip and Soler, Joan Ramon},
  booktitle={IJCAI 16},
  pages={766--772},
  year={2016}
}

@article{incMCQ,
  title={Incremental upper bound for the maximum clique problem},
  author={Li, Chu-Min and Fang, Zhiwen and Jiang, Hua and Xu, Ke},
  journal={INFORMS Journal on Computing},
  pages={137-153},
  year={2018},
  volume={30},
  number={1},
  publisher={INFORMS}
}

@inproceedings{bandmaxsat,
  title={Bandmaxsat: A local search MaxSAT solver with multi-armed bandit},
  author={Zheng, Jiongzhi and He, Kun and Zhou, Jianrong and Jin, Yan and Li, Chu-Min and Manyà, Felip},
  booktitle={IJCAI 22},
  pages={1901--1907},
  year={2022}
}

@inproceedings{aaai19Schedul,
  author       = {Junwen Ding and
                  Zhipeng L{\"{u}} and
                  Chu{-}Min Li and
                  Liji Shen and
                  Liping Xu and
                  Fred W. Glover},
  title        = {A Two-Individual Based Evolutionary Algorithm for the Flexible Job
                  Shop Scheduling Problem},
  booktitle    = {{AAAI} 2019},
  pages        = {2262--2271},
  publisher    = {{AAAI} Press},
  year         = {2019}
}

@inproceedings{aaai20MCS,
  title={A learning based branch and bound for maximum common subgraph related problems},
  author={Liu, Yanli and Li, Chu-Min and Jiang, Hua and He, Kun},
  booktitle={AAAI 20},
  pages={2392--2399},
  year={2020}
}

@article{gcplearning,
  title={An exact algorithm with learning for the graph coloring problem},
  author={Zhou, Zhaoyang and Li, Chu-Min and Huang, Chong and Xu, Ruchu},
  journal={Computers \& operations research},
  pages={282-301},
  year={2014},
  volume={51},
  publisher={Pergamon}
}

\end{document}